\documentclass{article}

\usepackage[margin=1in]{geometry} % Adjust margins as needed
\usepackage{setspace} % For line spacing adjustments
\usepackage{titlesec} % For formatting section titles
\usepackage{amsmath} % For mathematical formulas
\usepackage{graphicx}
\usepackage[colorlinks=true,linkcolor=blue,citecolor=blue,urlcolor=blue]{hyperref}
\usepackage{natbib} % Bibliography management
\setcitestyle{aysep={}} % This removes the comma between author and year
\usepackage{enumitem} % To control list environment
\usepackage{float}

\title{Classifying populist language in American presidential and governor speeches using automatic text analysis%
\thanks{Working paper. Van der Veen’s research has been partly funded by the European Union (ERC Consolidator grant, RADIUNCE, \#101043543). Views and opinions expressed are however those of the authors only and do not necessarily reflect those of the European Union or the European Research Council Executive Agency. Neither the European Union nor the granting authority can be held responsible for them.}
}

\author{Olaf van der Veen\footnotemark[1], Semir Dzebo\footnotemark[2], Levente Littvay\footnotemark[3], Kirk Hawkins\footnotemark[4], Oren Dar\footnotemark[5]}

\begin{document}

\maketitle

% Footnotes for Authors
\footnotetext[1]{Utrecht University, School of Governance.}
\footnotetext[2]{Central European University, Department of International Relations.}
\footnotetext[3]{HUN-REN Centre for Social Sciences, Institute for Political Science; Central European University, Democracy Institute.}
\footnotetext[4]{Brigham Young University, Department of Political Science.}
\footnotetext[5]{Independent Researcher.}
\date{} % Remove date

\doublespacing % Double-spacing
\pagestyle{plain} % Page numbers on all pages
\maketitle

\begin{abstract}

Populism is a concept that is often used but notoriously difficult to measure. Common qualitative measurements like holistic grading or content analysis require great amounts of time and labour, making it difficult to quickly scope out which politicians should be classified as populist and which should not, while quantitative methods show mixed results when it comes to classifying populist rhetoric. In this paper, we develop a pipeline to train and validate an automated classification model to estimate the use of populist language. We train models based on sentences that were identified as populist and pluralist in 300 US governors’ speeches from 2010 to 2018 and in 45 speeches of presidential candidates in 2016. We find that these models classify most speeches correctly, including 84\% of governor speeches and 89\% of presidential speeches. These results extend to different time periods (with 92\% accuracy on more recent American governors), different amounts of data (with as few as 70 training sentences per class achieving similar performance), and when classifying politicians instead of individual speeches. This pipeline is thus an effective tool that can optimize the systematic and swift classification of the use of populist language in politicians’ speeches.

\end{abstract}

\newpage

\section{Introduction}

In 2017, Cambridge University Press named populism their 'word of the year'. In the wake of the successful Brexit campaign and Trump's election as American president, populism became an important way to understand these unfolding events. In the past seven years, the concept has remained important, with populist leaders elected in many advanced democracies, and with populists often testing or eroding checks and balances when elected \citep{Meyer_2024}. As such, the measurement of populist rhetoric has remained paramount to a complete understanding of the current political landscape.

This measurement, however, has not always proved easy. Methods commonly used to gauge if political representatives use populist language - such as holistic grading \citep{hawkinsChavezPopulistMeasuring2009} or content analysis \citep{rooduijnMeasuringPopulismComparing2011} - are very labour-intensive, often requiring several coders to evaluate the same speech. As a result, full coverage of a system is uncommon, with coding often focusing on parties or politicians that were already considered likely populists (e.g. \citealt{Franzmann_Lewandowsky_2020}; exceptions include \citealt{Dzebo_Jenne_Littvay_Hawkins_Veen_2024}; \citealt{hawkinsGlobalPopulismDatabase2022}). On the other hand, methods such as expert coding often get used for the evaluation of parties' populism through static means like manifestos (e.g. \citealt{rooduijnMeasuringPopulismComparing2011}). They may therefore not accurately gauge individual representatives who can use differing amounts of populist language depending on context.

To combat this, scholars have increasingly employed computational methods to identify and analyze populist rhetoric in political discourse (e.g. \citealt{di_cocco_how_2022}; \citealt{erhard_popbert_2023}). These methods, if effective, may save a large amount of time and labour, allowing scholars to classify many more politicians. This would allow researchers to more easily achieve full coverage of a oolitical system, rather than only looking at likely populists or large parties. However, these methods have for now been used to classify sentences (\citealt{erhard_popbert_2023}; \citealt{klamm_our_2023}) or party manifestos \citep{di_cocco_how_2022}, not individual speeches or politicians.

In this paper, we develop a pipeline (i.e. process) to train and validate an automated classification model which estimates the use of populist language. We use this pipeline to train models using speeches of various American politicians, including governors and presidential candidates. We show that is pipeline is effective at creating models that classify populist rhetoric in these politicians. It works effectively in a variety of contexts, at both the speech- and politician-level, and with small amounts of training data. We also determine the boundary conditions of these models, showing where exactly its effectiveness drops off. In doing so, we contribute to the body of literature that explores the uses and limits of large language models in social science \citep{baden_three_2022}. We argue that the pipeline presented in this paper can become more widely used for the classification of populist language in political speeches, saving great amounts of time and labour.

This paper is structured as follows. First, we discuss the concept of populism, and the ways in which populist language is suitable for empirical analysis via computational approaches. Second, we introduce the pipeline we use to fine-tune a large language model so that it can find populist language, and the datasets we use to train and validate this pipeline. We then present the results, showing that the models created through this pipeline are effective at detecting populism at the sentence-, speech-, and speaker-level. We end by discussing the implications of this.

\section{Literature review}

\subsection{Populism as Ideational Discourse}

The concept of populism has gained significant attention in political science research, with various approaches proposed for defining and measuring this complex phenomenon. These approaches include the ideational approach, which conceptualizes populism as a set of ideas emphasizing a Manichean struggle between the virtuous ``people" and the corrupt ``elite" \citep{hawkinsIntroductionIdeationalApproach2018}; the strategic approach, which describes populism an individual's strategic use of a large number of followers to gain political power (\citealt{Weyland_2001}, p. 12); and the performative or stylistic approach, which views populism as a political style that emphasises ``the people" over ``the elite", frequently contains bad manners, and often focuses on crises (\citealt{Moffitt_2016}, p. 45). 

Of these, the ideational approach is most appropriate for this study. The ideational definition of populism, pioneered by scholars such as Cas Mudde (\citeyear{muddePopulistZeitgeist2004}), considers populism as a ``thin-centered ideology" that can attach itself to diverse political platforms. It focuses on identifying the core populist ideas expressed in political discourse, rather than associating populism with specific policy positions or political strategies. This flexibility allows populism to manifest in both left-wing and right-wing variants, as it provides a discursive framework for interpreting political reality rather than a comprehensive normative worldview \citep{muddeExclusionaryVsInclusionary2013}. At the same time, the clearly delineated definition allows it to be easily  operationalised, and has therefore become most widely used for empirical analysis \citep{Meijers_Zaslove_2021}. 

The ideational approach's emphasis on populism as a set of ideas aligns closely with the concept of discursive framing in political communication research. Discursive frames are interpretive schemas that shape how individuals perceive and make sense of political issues \citep{goffmanFrameAnalysisEssay1974, snowFrameAlignmentProcesses1986, chongTheoryFramingOpinion2007}. Political actors strategically employ frames to promote their preferred interpretation of reality and mobilize support for their positions. Populist ideas, when expressed in political discourse, serve as a distinct discursive frame \citep{caianiElitistPopulismExtreme2011}. The populist frame is characterized by a valorization of ``the people", a denigration of ``the elite", and a portrayal of politics as a moral struggle between these two groups. By analyzing the presence and intensity of populist framing in political texts, researchers can measure the degree of populist ideation employed by political actors.

This conceptualization of populism as a discursive phenomenon, rooted in the ideational approach, provides a foundation for systematic assessment of populist rhetoric. Researchers can examine political discourse through this lens to evaluate the extent to which political actors employ populist ideas. Such an approach facilitates the application of various methodological techniques to study populism as a framing strategy in political communication.

\subsection{Computational Approaches to Measuring Populist Discourse}

The ideational approach's operationalizable focus on populism as a discursive phenomenon has motivated the development of systematic methods for measuring populist ideas in political texts. Early studies relied on labour-intensive hand-coding by trained experts to identify populist discourse \citep{hawkinsChavezPopulistMeasuring2009, rooduijnPopulistZeitgeistProgrammatic2014, hawkinsGlobalPopulismDatabase2022, jenneMappingPopulismNationalism2021}. However, the growing availability of large-scale digital text data has spurred the development of computational methods for automated populist discourse analysis.

Existing computational approaches to measuring populist discourse include dictionary-based approaches, topic modeling, network-based methods, word embeddings, and fine-tuning large language models. In dictionary-based analysis, researchers use predefined lists of populist keywords to measure the frequency of populist language in texts \citep{rooduijnMeasuringPopulismComparing2011}. Topic modeling techniques, such as Latent Dirichlet Allocation (LDA) \citep{bleiLatentDirichletAllocation2003}, have been used to identify latent themes and topics associated with populist discourse in political text \citep{mullerRightWingPopulistControlled2022}. Network-based approaches have been employed to analyze the co-occurrence of populist terms and phrases, revealing the underlying structure and relationships within populist discourse \citep{fernandezgarciaPopulismPeopleAnalysis2020, newthCommonSensePopulism2023}. Word embeddings, which represent words as dense vectors in a high-dimensional space, have been used to capture the semantic similarities and relationships between populist and non-populist terms \citep{daiWhenPoliticiansUse2022}.

While these approaches have contributed to our understanding of populist discourse, they have limitations in capturing the nuanced and context-dependent nature of populist expressions. Dictionary-methods rely on predefined keywords lists, which can miss the subtle ways in which populist ideas are articulated. Topic modeling and network-based approaches provide insights into the themes and relationships within populist discourse but may not directly identify populist content at the sentence level. Word embedding may capture semantic similarities, but creating accurate embeddings from scratch requires large amounts of data. 

Fine-tuning a large language model (LLM) offers a more nuanced approach because it combines the strong baseline understanding of LLM with the contextual information provided by human-coded examples. This method takes an existing LLM as a baseline, and provides additional human-coded input, which allows the model to learn the textual patterns in that specific context. The use of a strong baseline model theoretically allows for effective performance even with less training data. A recent study by Di Cocco and Monechi (\citeyear{di_cocco_how_2022}) applied this method to measure populism in party manifestos. However, their approach has been critiqued for its choice of training data. The study used full manifestos as input data, coding all sentences from populist parties as populist. In response, Jankowski and Huber (\citeyear{jankowskiWhenCorrelationNot2023}) argued that this approach introduces bias, as the models can rely on a wide range on concepts beyond populism to predict whether a setence comes from a populist party. They emphasize that not every sentence from a populist party is inherently populist.

Our study addresses this methodological challenge by providing an existing LLM with a carefully curated dataset of sentences that have been coded as either populist or non-populist. By focusing on these clear examples of populist discourse, we aim to improve the validity of automated fine-tuning approaches. Our model learns to identify the specific textual features associated with populist ideas, rather than relying on party-level labels or other potentially confounding factors. This targeted training data allows our model to more accurately capture the subtleties of populist discourse in political speeches. 

\section{Data and Methods}

As described above, our study fine-tunes an LLM by providing it a variety of populist and non-populist sentences as training data to improve its contextual understanding. We then use the fine-tuned model to classify speeches by having it classify all sentences in a speech. The percentage of the sentences in a speech classified by the model as populist reveals the pervasiveness of populist language in that speech. To assess the model's validity, its classifications are compared against expert human coding. Below, we elaborate on each step.

\subsection{Baseline model}

As a baseline model, we rely on a transformer model similar to  BERT (Bidirectional Encoder Representations from Transformers; \citealt{devlin_bert_2019}), which is a pre-trained model that has gained widespread use in the social sciences (e.g. \citealt{bosley_we_2023}; \citealt{shen2022sscibert}; \citealt{erhard_popbert_2023}). Given a natural language input (such as a sentence), the model produces  so-called embeddings as outputs, which are mathematical representations that capture the meanings of words in context. BERT’s training on large bodies of text data has given it a nuanced baseline understanding of language. In particular, we rely on the Sentence-transformers framework (SBERT; \citealt{reimers2019sentencebert}), which creates embeddings at the sentence-level rather than for each word (for example by averaging all the embeddings), allowing for faster performance when analysing full sentences while still retaining contextual information. This strong baseline understanding can then be fine-tuned by providing high-quality training data on the concept of interest, enabling the model to effectively adapt to the particular context.

\subsection{Data}

For fine-tuning and prediction, we use three separate datasets. The first comes from \citet{Dzebo_Jenne_Littvay_Hawkins_Veen_2024}, who score the use of populist language in speeches from various American governors with terms between 2012 and 2020. For each governor term, four speeches are scored: (1) a campaign speech; (2) a state-of-the-state speech; (3) a ribbon-cutting (ceremonial) speech; and (4) a famous speech. The speeches are all scored on a scale of 0 (no populism at all) to 2 (strong populism throughout), up to a single decimal point. This is done using holistic grading, with coders grading a speech based on their holistic assessment and providing justification for their grading in coding rubrics. During processing, some speeches were not imported correctly; only terms that contained at least three speeches were kept so that scores for different governors could be compared. All in all, this means the dataset contains 288 speeches in total, coming from 73 terms in 37 distinct states. 

In addition, we rely on coded speeches from presidential candidates in 2016, for which we used the dataset and rubrics from \citet{Hawkins_2016}, discussed in Hawkins and Littvay (\citeyear{hawkins_contemporary_2019}). This dataset contains 45 usable speeches in total. The number of speeches coded was not evenly distributed among the seven candidates coded: the eventual nominated candidates, Donald Trump and Hillary Clinton, both had 16 speeches in this dataset, whilst other candidates had four coded speeches at most (Bernie Sanders 4; Ted Cruz and Marco Rubio 3; John Kasich 2; and Ben Carson 1). Since these speeches were from the presidential campaign, they are all of the same type (namely `campaign’ speeches). Speeches were coded using the same holistic grading method as described above.

Both datasets provide sentences which we can use to train the model. During the holistic grading process, coders were instructed to provide justification for their scores in rubrics. This justification included passages which the coders found particularly indicative of either populism or pluralism (considered as the opposite of populism in that study). We collect these passages and split them into sentences. Because a real speech contains fairly few populist sentences – at most 15\% for Party Manifestos, according to \citet{rooduijnMeasuringPopulismComparing2011}, with speeches likely containing no more than double (\citealt{Stuvland_2021}, p. 78) – and the same likely applies to pluralism, we add a third `neutral’ category of sentences to our training data. To enhance the model's discriminatory capacity, we systematically reviewed all speeches and extracted passages that exemplified neither populist nor pluralist rhetoric (see Appendix 4 for examples across all categories). These sentences combined provide the training data on the basis of which a model can learn to distinguish between populist and non-populist sentences. For the governor data, this created a dataset of 1400 populist, 1009 pluralist, and 1121 neutral sentences. For the presidential candidates, we ended with 326 populist sentences, 148 pluralist sentences, and 140 neutral sentences. Both of these form the foundation for the fine-tuning of two separate models.

In addition, to test the model trained on governor data, we import an additional dataset of six governors from 2018 to 2022, coded using the same method as described above. Once again, four speeches were coded for each governor, with a fifth for a single governor, giving us 25 speeches in total. This dataset allows us to evaluate model performance for a different time period.

\subsection{Fine-tuning}

The fine-tuning is done using SetFit (Sentence Transformer Fine-Tuning; \citealt{tunstall_efficient_2022}), a pipeline which fine-tunes an already-strong baseline model based on these embeddings to find common characteristics for each category (in this case: populist, pluralist, and neutral). SetFit allows for good performance even with relatively little training data, increasing the applicability of the model. We tried several versions of the model, each with slightly different parameters, eventually selecting the most accurate one (see Appendix 1). Many versions of the model performed similarly, indicating good robustness. We ran the model with a single epoch and a batch size of 6 for computational efficiency. All versions of the model ran with a 75/25 train/test split, meaning 75\% of the sentences were used to train the model, and 25\% were used to check its performance.

For preprocessing, both the training passages and the speeches were split into sentences by punctuation mark. Additional preprocessing steps (such as changing all words to lowercase, stemming words, and removing stop words), although generally important for automatic text analysis \citep{Denny_Spirling_2018}, are not appropriate for our method, which incorporates contextual information. As such, these steps were not taken.

One potential threat for the validity of the above method is `data leakage' - namely that the training sentences on which the model fine-tunes might come from the same speeches on which the model later classifies. If unchecked, this would mean the model scores sentences it has seen before. To address this, we maintain a record of each sentence's original speech context. This was either done through preprocessing or by matching sentences post-hoc. The function we created was effective for catching most passages, mitigating this potential issue (see Appendix 2 for more discussion).

After matching, we train the model separately for each term or speaker, making sure to exclude matching passages from the training data. This ensures that the model never scores sentences which it also trained on. As a result, the model is trained 73 times for the governors, and 7 times for the presidential candidates. After training, the model classifies all sentences in a given speech (again split by punctuation mark) as populist, pluralist, or neutral. 

\subsection{Prediction}

For predicting the use of populist rhetoric in speeches and by speakers, we employ a binary classifier (populist or non-populist) rather than the continuous 0-2 scale used in the original datasets. In both the governor and the presidential data, only few speeches were classified as very populist (1.5 or higher), leading to situations of data sparsity at those levels. By changing it to a binary classifier, we ensure that there are ample speeches in both categories. We chose 0.5 as the cut-off, as this means that there is the presence of at least some populist language \citep{Hawkins_Castanho_Silva_2016}. This modification alters the substantive focus of the model, shifting from measuring the prevalence of populist language to detecting its presence. While this approach is methodologically more appropriate, we also provide model performance results using the original continuous variables in Appendix 3.

To classify speeches, we only use the percentage of speeches that the model deemed populist. We found that adding the percentage of sentences classified as pluralist had no impact on performance, and thus decided not to use it in classification. Because this means speeches are classified based only on a single parameter, a simple decision model is more appropriate than a more complex model like a Random Forest algorithm (used in e.g. \citealt{di_cocco_how_2022}). Therefore, we use a decision tree classifier with the maximum number of decisions as 1, meaning it chooses only a single cutoff. To show the models expected results, results below show the modal value of 100 runs of a decision tree.

This same process was applied to test the effectiveness of the model at classifying speakers (governors or presidential candidates). Speakers were classified as populist if their scores averaged to more than 0.5; the model considered the percentage of all sentences made by a speaker that were classified as populist, and classified the speaker as populist if this percentage exceeded the cutoff set by the decision tree.

\section{Results}

\subsection{Model validity}

We first test the model accuracy at the sentence-level. For completeness, we compute several metrics which score model effectiveness. Precision, recall, and F1 are all commonly used metrics for classification tasks. Precision determines the proportion of predicted positives that are actually positive (here, populist); recall the proportion of true positives that were predicted as positives (percentage of populist sentences correctly predicted); F1 combines the two scores for a more general judgment. Where appropriate, we also compute the area under the ROC-curve (AuROC) and Matthews Correlation Coefficient (MCC) as two additional metrics that consider both false positive and false negative rates. All metrics range from 0 to 1, where 0 usually indicates either randomness or complete failure and 1 indicates perfect predictive power.

The model achieves a 70 \% accuracy for the presidential data and a slightly lower 66\% accuracy for the governor data, with F1 scores of 0.75 and 0.71 respectively. The full statistics can be found in table 1. This is much better than random (which would be 33 \%, since there are three roughly evenly sized categories), and sits between F1 scores reported by \citet{di_cocco_how_2022} and those reported by \citet{erhard_popbert_2023}. This aligns with expectations, as our training data is more fine-grained than the data from \citet{di_cocco_how_2022}, who train based on entire manifestos where the populist content is thus more loosely spread, but slightly less fine-grained than that of \citet{erhard_popbert_2023}, who annotate individual sentences rather than groups of sentences through a much more labour-intensive process.

\begin{table}[H]
    %\centering
    \begin{tabular}{|c|c|c|c|c|c|c|}
        \hline
         \textbf{Model}  & Accuracy (std) & Precision (std)  & Recall (std)  & f1 (std) & F2 (std) & MCC (std) \\
         \hline
         Governors & 0.66 & 0.73 & 0.69 & 0.71 & 0.70 & 0.50  \\
         \hline
         Presidential candidates & 0.70 (0.03) & 0.77 (0.06) & 0.74 (0.06) & 0.75 (0.06) & 0.55 (0.04) \\
         \hline
    \end{tabular}
    \caption{Hyperparameter tuning results}
    \label{tab:my_label}
\end{table}

We ensure that the model detects populism - as opposed to related concepts like combativeness or other rhetoric strategies - by examining the most erroneously classified speeches. This analysis shows that, for all three datasets, the speeches with the largest discrepancy between predicted and actual populism were speeches that contain a high degree of combativeness against the speaker's opponent - California governor Gavin Newsom opposing Republican contender Larry Elder and Minnesota governor Mark Dayton and presidential candidate Hillary Clinton both opposing Donald Trump. Intuitively, this makes sense, because populism's black-and-white worldview often contains combative elements. Still, the two concepts also differ, and it would be concerning if the model did not distinguish between them. 

Further analysis, however, eases concerns that our measure of populism is only a proxy for combativeness. For one, comparing our measure with measures of combativeness - such as a 'divisiveness' dictionary developed by \citet{zhou_quantifying_2024} - shows moderate but not extreme agreement between the two measures $(r^2 \text{ of } 0.28\text{--}0.3)$, roughly as would be expected of two tangentially related concepts. Further qualitative analysis reveals numerous speeches that were divisive yet correctly classified as non-populist by the model—such as another speech from Hillary Clinton opposing Donald Trump. Conversely, the model accurately identified speeches employing populist rhetoric without divisive language, exemplified by a speech from Louisiana Governor Bobby Jindal. These speeches are included in Appendix 5.

\subsection{Model performance}

The model trained on presidential data performs well. It predicts six of the seven presidential candidates correctly, including three presidential candidates that use populist language (Carson, Trump, and Sanders. It only misses Ted Cruz. It also predicts 40 out of the 45 speeches correctly, which includes identifying 17 out of 20 speeches with populist rhetoric (AuROC 0.89; F1 0.87; MCC 0.77). Figure 1 shows these results.

\begin{figure}[H]
    \includegraphics[width=0.4\linewidth]{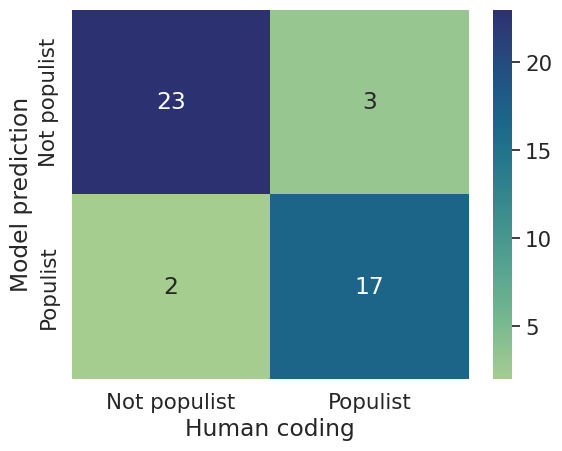}
    \includegraphics[width=0.4\linewidth]{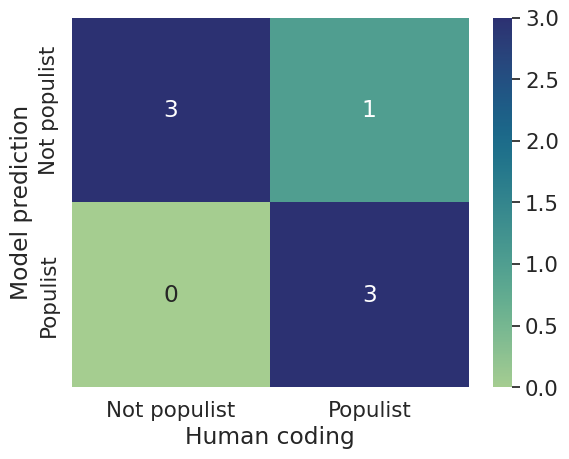}
    \caption{Confusion matrices for presidential candidates, speeches (left) and speakers (right)}
    \label{fig:enter-label}
\end{figure}

The model trained on governor data performs worse when evaluating speeches of governors with 2010-2018 terms. It correctly predicts 62 out of 77 governors, but identifies only 10 out of 17 populist governors (AuROC 0.77; F1 0.69; MCC 0.60). Performance stays modest on the term level, classifying 244 of 288 speeches correctly, and identifying 48 of the 60 populist speeches (AuROC 0.72; F1 0.57; MCC 0.45). These results can be found in figure 2. 

\begin{figure}[H]
    \includegraphics[width=0.4\linewidth]{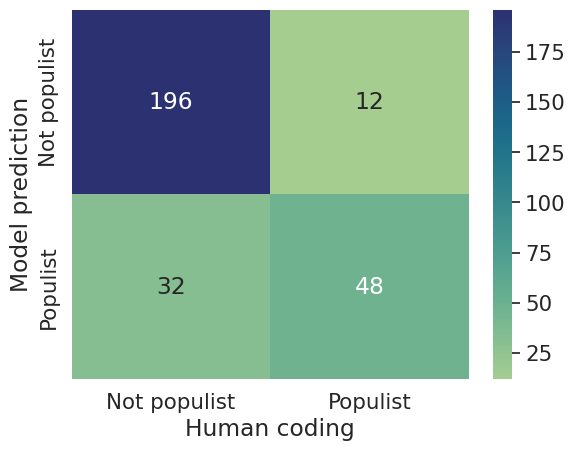}
    \includegraphics[width=0.4\linewidth]{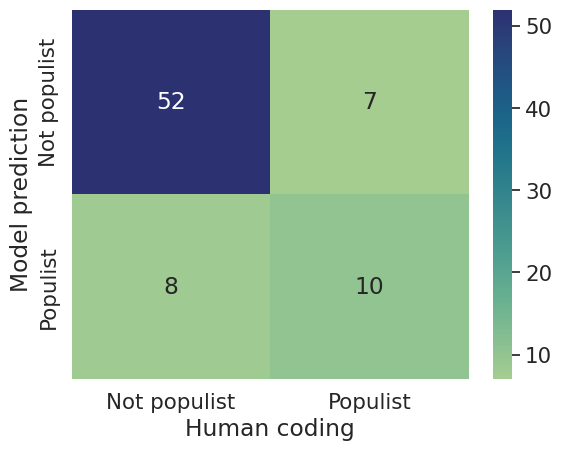}
    \caption{Confusion matrices for governors with 2010-2018 terms, speeches (left) and speakers (right)}
    \label{fig:enter-label}
\end{figure}

Finally, the model performs well on the more recent governor data. It correctly predicts five of the six governors, correctly identifying DeSantis as populist and but missing Noem as populist. At the speech level, it correctly classifies 24 out of 26 speeches and identifies all ten populist speeches (AuROC 0.89; F1 0.8; MCC 0.71). Figure 3 displays these results.

\begin{figure}[H]
    \includegraphics[width=0.4\linewidth]{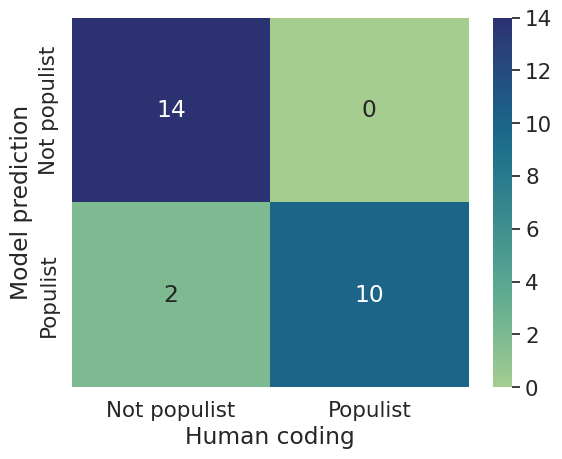}
    \includegraphics[width=0.4\linewidth]{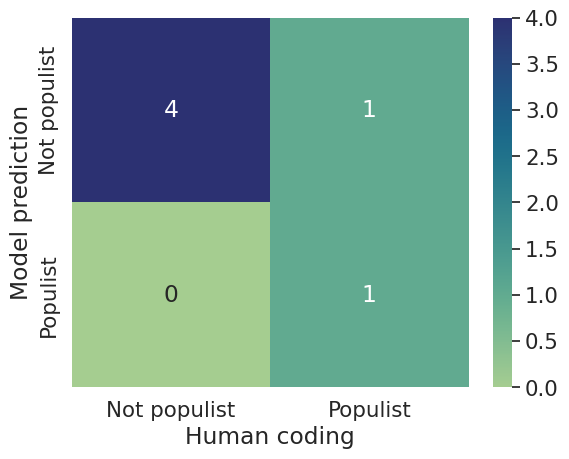}
    \caption{Confusion matrices for governors with 2018-2022 terms, speeches (left) and speakers (right)}
    \label{fig:enter-label}
\end{figure}

Table 2 shows the relevant statistics for all three datasets. In addition to the statistics provided above, we also include the 'F2'-score. Like F1, this combines precision and recall into a holistic judgment. However, it weighs recall more highly, meaning biases towards punishing 'false negatives'. Here, the F2 score thus more heavily penalizes false negatives—populist speeches not classified as populist. This metric aligns with the model's primary purpose of identifying populist rhetoric in political discourse. 

\begin{table}[H]
    \centering
    \begin{tabular}{|c|c|c|c|c|c|c|c|c|}
        \hline
        Data & N & Accuracy & Precision & Recall & F1 & F2 & AuROC & MCC  \\
        \hline
        Governors 2010-2018 speeches & 288 & 0.85 & 0.8 & 0.6 & 0.69 & 0.63 & 0.77 & 0.70 \\
        \hline
        Governors 2018-2022 speeches & 26 & 0.92 & 1 & 0082 & 0.9 & 0.85 & 0.91 & 0.85 \\
        \hline
        Presidential candidates speeches & 45 & 0.89 & 0.85 & 0.89 & 0.87 & 0.89 & 0.89 & 0.77 \\
        \hline
        Governors 2010-2018 speakers & 77 & 0.81 & 0.59 & 0.56 & 0.57 & 0.56 & 0.72 & 0.45  \\
        \hline
        Governors 2018-2022 speakers & 6 & 0.83 & 0.5 & 1 & 0.67 & 0.83 & 0.9 & 0.63   \\
        \hline
        Presidential candidates speakers & 7 & 0.86 & 0.75 & 1 & 0.86 & 0.93 & 0.88 & 0.75 \\
        \hline
    \end{tabular}
    \caption{Model results for speeches}
    \label{tab:my_label}
\end{table}

\subsection{Boundary conditions}

\subsubsection{Impact of context}

To investigate the impact of context, we test the models in the opposite context. In other words, we test the model trained on governor sentences on the speeches of presidential candidates, and vice versa. First, we look at whether the general level of inferred populism changes. We find that, indeed, the proportion of sentences that gets graded as populist depends greatly on the context. When the model trained on governor data classifies the governor speeches, it classifies 21.9\% of sentences as populist; this drops to 15.9\% when the model trained on presidential data classifies the governors. Conversely, when the model trained on presidential data classifies presidential speeches, it classifies 21.6\% of sentences as populist; this increases to a staggering 46.8\% when the governor data predicts it. This indicates that the training sentences from the presidential context were likely more extreme and less subtle than those in the governor context, implying that governors may use less extreme language when expressing populism.

Even though the number of sentences classified as populist changes greatly, performance remains strong for all models. The model trained on presidential data achieves similar performance to even the model trained on governor data despite being adapted to a different context, with an accuracy of 86\% and catching 43 out of 60 populist speeches (AuROC 0.81; F1 0.68; MCC 0.60). The model trained on governor data performs worse on presidential data, with an accuracy of 0.71 and catching 16 out of 20 populist speeches (AuROC 0.72; F1 0.71; MCC 0.44). 

In addition, we investigate the impact of speech type by testing the model trained on presidential candidates on campaign speeches of governors. Since all speeches made during candidacy are by nature campaign speeches, this aligns the speech type of these speeches. Thus, we would expect the performance of the model to improve further. This turns out correct, although differences are minor and could have resulted from chance: AuROC, F1, and MCC scores all improve by between 0.00 and 0.08 (AuROC 0.81; F1 0.76; MCC 0.62). All metrics can be found in table 3.

\begin{table}[H]
    \centering
    \begin{tabular}{|c|c|c|c|c|c|c|c|c|}
        \hline
        Data & N & Accuracy & Precision & Recall & F1 & F2 & AuROC & MCC  \\
        \hline
        Presidential training; governor testing & 288 & 0.86 & 0.65 & 0.72 & 0.68 & 0.70 & 0.81 & 0.60 \\
        \hline
        Presidential training; governor campaign testing & 71 & 0.82 & 0.72 & 0.81 & 0.76 & 0.79 & 0.81 & 0.62 \\
        \hline
        Governor training; presidential testing & 45 & 0.71 & 0.64 & 0.8 & 0.71 & 0.76 & 0.72 & 0.44 \\
        \hline
    \end{tabular}
    \caption{Model results in different contexts}
    \label{tab:my_label}
\end{table}

In conclusion, these results indicate that context did not have a large impact on model performance in this situation. The predictiveness of the training sentences - as indicated by the difference in accuracy of the two models at the sentence level - seemed to have a greater influence on performance. 

\subsubsection{Impact of data sparsity}

For the presidential data, the imbalanced nature of the data (with Trump and Clinton appearing more frequently than other candidates), combined with the matching function removing sentences for speakers when their speeches contains those sentences, means that Trump and especially Clinton have fewer training sentences than the other candidates (337 and 227 respectively, compared to 400 or more for other candidates). This allows us to gauge the impact of data sparsity on model performance. For this data, we see that model performance remains stable for both candidates, with accuracy for Clinton remaining at 68\%, indicating that as few as 75 training sentences per class may be enough for good performance.

To find out more, we run the model on the governor data several times, giving it an artificially reduced number of sentences per category each time. The results of this can be found in table 4 and figure 1. The model shows a steep drop-off in performance when fewer than 70 training sentences per category are available, dropping in accuracy by 20\%, in F1-score by 0.17, and in MCC by 0.29. This gives an early indication of the minimum number of training sentences required for sufficient performance in this situation.

\begin{table}[H]
\centering
\begin{tabular}{|c|c|c|c|c|c|}
\hline
Number of sentences per class & Accuracy & Precision & Recall & F1  & MCC  \\ \hline
1000 & 0.66     & 0.73      & 0.69   & 0.71 & 0.5  \\ \hline
400  & 0.62     & 0.71      & 0.67   & 0.69 & 0.43 \\ \hline
250  & 0.58     & 0.62      & 0.6    & 0.61 & 0.37 \\ \hline
150  & 0.6      & 0.61      & 0.59   & 0.6  & 0.41 \\ \hline
100  & 0.53     & 0.57      & 0.55   & 0.56 & 0.29 \\ \hline
90   & 0.61     & 0.8       & 0.55   & 0.65 & 0.43 \\ \hline
80   & 0.61     & 0.74      & 0.58   & 0.65 & 0.42 \\ \hline
70   & 0.64     & 0.62      & 0.59   & 0.61 & 0.46 \\ \hline
60   & 0.44     & 0.46      & 0.4    & 0.43 & 0.17 \\ \hline
50   & 0.43     & 0.36      & 0.3    & 0.33 & 0.15 \\ \hline
40   & 0.5      & 0.62      & 0.4    & 0.48 & 0.28 \\ \hline
30   & 0.43     & 0.42      & 0.71   & 0.53 & 0.17 \\ \hline
\end{tabular}
\caption{Model results for different numbers of sentences}
\label{tab:my_label}
\end{table}

\begin{figure}[H]
    \centering
    \includegraphics[width=0.32\linewidth]{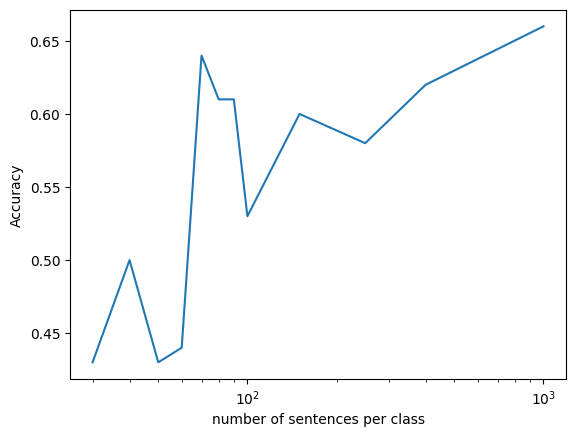}
    \includegraphics[width=0.32\linewidth]{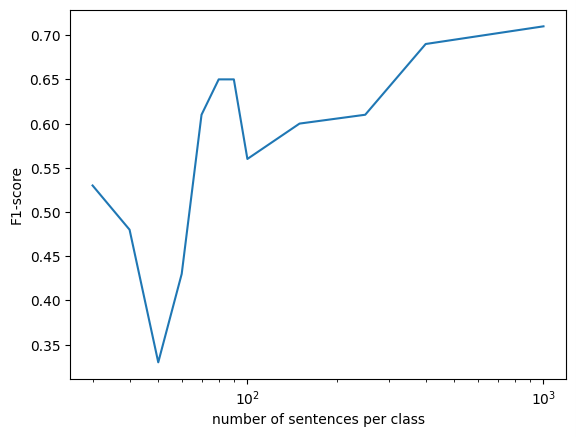}
    \includegraphics[width=0.32\linewidth]{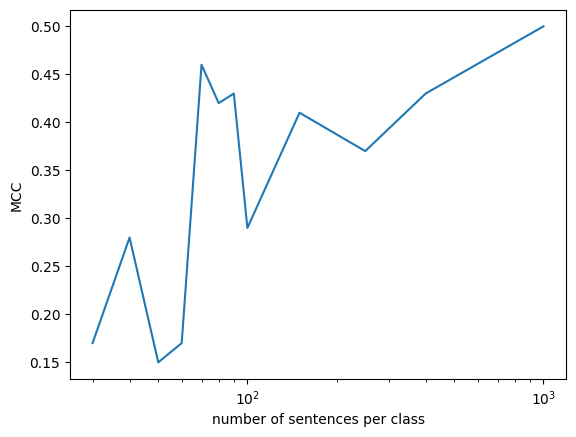}    
    \caption{Model performance by number of sentences per class}
    \label{fig:enter-label}
\end{figure}

\section{Discussion and Conclusion}

In short, the pipeline presented above creates well-performing models that accurately capture populist rhetoric in several contexts and given many different starting parameters. Encouragingly, the model performs well with as few as 70 to 100 training sentences per category. This means it could likely be used in a variety of situations, once that number of sentences has been collected. This would allow researchers to much more easily get full coverage of a particular political system, which is usually very labour-intensive. In particular, this more easily allows estimations on populist rhetoric in all political representatives, rather than focusing only on parties or executives.

Furthermore, the model could be changed to improve performance further. For example, our model only runs on a single epoch now to save on computational expense. Running for multiple epochs or with a larger batch would likely improve performance. In addition, we ignore sentence order in our current model, which scholars in political communication have identified as a potentially relevant feature for political phenomena (e.g. \citealt{baden_three_2022}, p. 4). Future research could further refine the model by including these factors. The strong performance of this model despite the potential for further improvements, alongside the quick development of LLMs, bodes well for the future of automatic text analysis in social science.

It runs counter to prior expectations that the model finds little impact of pluralism on the judgment of the speech as populist. Initial research on holistic grading took this juxtaposition from the discursive literature \citep{hawkinsChavezPopulistMeasuring2009}. This research questions the value of this contrast, suggesting that perhaps a label of simply 'non-populist' suffices.

The equivalent performance of the models in different contexts also runs counter to prior expectations. Context differences are generally believed to play an important role in the expression of social concepts, including differences in speaking platform \citep{Schoonvelde_Schumacher_Bakker_2019} and differences between countries \citep{Baden_Dolinsky_Lind_Pipal_Schoonvelde_Shababo_van_der_Velden_2022}. The analysis in this paper indicates that the expression of populism does not change between the sub-national and national levels in the United States. Further analysis will need to show whether this results from the similarity of the two contexts (as both still pertained to the same political system), the concept of populism being uniquely stable, or the general principle of context-instability not being as iron-clad as previously believed. 

As discussed above, error analysis shows that the model still struggles at times with the distinction between populism (which requires a Manichean worldview applied to the 'people' and the elite) and broader combativeness (which contains this Manichean element, but not necessarily the distinction between people and elite). Future research may want to separate populism down into its subdimensions and have an automatic text analysis model classify each (following e.g. \citealt{klamm_our_2023}). This is especially useful since populism requires all elements \citep{WUTTKE_SCHIMPF_SCHOEN_2020}. The coding rubrics provided in the datasets used for this study (e.g. \citealt{Dzebo_Jenne_Littvay_Hawkins_Veen_2024}) usually include separate sections for these elements. This can thus be implemented using a similar pipeline to the one used in this study. 

\newpage
% Bibliography
\bibliographystyle{chicago} % 
\bibliography{Populism_AutomatedSpeechCoding}

\newpage
\section{Appendix}
\subsection{Hyperparameter tuning}

We tried 3 different hyperparameters in addition to the default model. Firstly, we tried adding  `differential heads', which allows the model to attend to different features of the language simultaneously. Secondly, we tried `end-to-end training', which allows the model to update a wider array of parameters when fine-tuning. Finally, we tried a different embedding model than SBERT, called BGE-M3 (Chen et al., 2024). We also tried all combinations of these changes. We tested all versions on the presidential data, because it would be too computationally expensive to test this many versions of the model on the governor data because of the larger dataset. In each case, we trained the model 10 times. In the end, the end-to-end training by itself performed the best. See table 5 below for the full results, with the mean and standard deviation for each setup over the 10 runs. 

Based on these results, the model seems to perform well in many different configurations - including the default configuration. This means it is likely robust and not dependent on the particular hyperparameters we chose.

\begin{table}[H]
    %\centering
    \begin{tabular}{ccccccc}
         \textbf{Model}  & Average accuracy (std) & Precision (std)  & Recall (std)  & F1 (std) & MCC (std) \\
         Baseline (1) & 0.69 (0.04) & 0.79 (0.05) & 0.72 (0.06) & 0.75 (0.05) & 0.54 (0.06)  \\
         Differential head (2) & 0.70 (0.04) & 0.76 (0.06) & 0.75 (0.05) & 0.75 (0.06) & 0.55 (0.06) \\
         Newer model (3) & 0.66 (0.02) & 0.75 (0.06) & 0.67 (0.03) & 0.71 (0.05) & 0.49 (0.03)\\
         End-to-end (4) & 0.70 (0.03) & 0.77 (0.06) & 0.74 (0.06) & 0.75 (0.06) & 0.55 (0.04) \\
         2 + 3 & 0.64 (0.05) & 0.69 (0.07) & 0.67 (0.03) & 0.68 (0.05) & 0.47 (0.07) \\
        2 + 4 & 0.65 (0.05) & 0.72 (0.07) & 0.68 (0.05) & 0.70 (0.06) & 0.48 (0.08) \\
         3 + 4 & 0.68 (0.03) & 0.72 (0.06) & 0.7 (0.09) & 0.71 (0.07) & 0.52 (0.04) \\
         2 + 3 + 4 & 0.63 (0.03) & 0.70 (0.12) & 0.65 (0.08) & 0.67 (0.09) & 0.44 (0.05) \\
    \end{tabular}
    \caption{Hyperparameter tuning results}
    \label{tab:my_label}
\end{table}

\subsection{Discussion of matching}

To avoid data leakage, the training sentences had to be matched to the speeches in which they were spoken. The populist training sentences were always matched to the speeches in preprocessing, meaning they all matched. The same goes for the pluralist sentences of the presidential dataset. 

For the governor data, this means that the pluralist and neutral sentences had to be matched post-hoc. We did this using a function that looked at each sentence and tried to match it to the speech where it appeared. This worked in most cases (though not all, because coders had sometimes changed the language to make it gramatically correct). Specifically, 73\% of pluralist and 82\% of neutral sentences found a match to a speech. However, this does not account for speeches which were lost during pre-processing (about 11.4 \% of the sample), the training sentences of which therefore cannot find a match. In reality, this means that about 83\% and 93\% of pluralist and neutral sentences respectively found a matching speech. For the neutral data of the presidential dataset, this number was 79 \%. 

Ensuring that this did not influence performance was one of the primary reasons why we also included analysis on the more recent governors, the sentences of which the model had definitely never seen. The stellar performance on this data indeed indicates that our function sufficiently excluded matches.

\subsection{Performance as continuous variable}

Below, in figures 1-3, we show the scatter plots that plot the correspondence between the proportion of sentences predicted by the model as populist (x-axis) and the score given by human coders (y-axis). In all cases, there was at least moderate correlation between the two scores. In each case, correlation was higher for speakers, when compared to speeches. For the presidential candidates, r2 was 0.58 for speeches and 0.62 for speakers; for the governors with terms from 2010-2018, r2 was 0.39 for speeches and 0.42 for speakers; for the governors with terms from 2018-2022, r2 was 0.49 for speeches and 0.74 for speakers

\begin{figure}
    \includegraphics[width=0.5\linewidth]{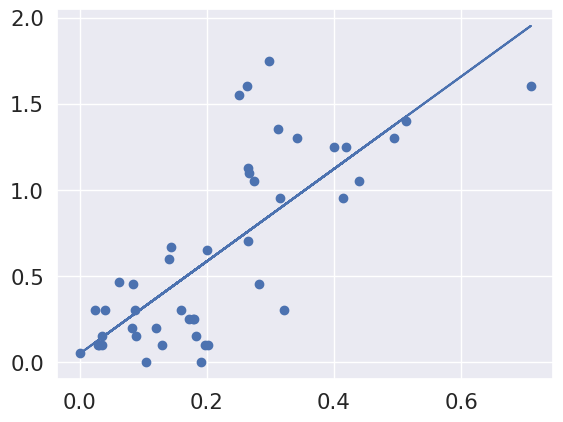}
    \includegraphics[width=0.5\linewidth]{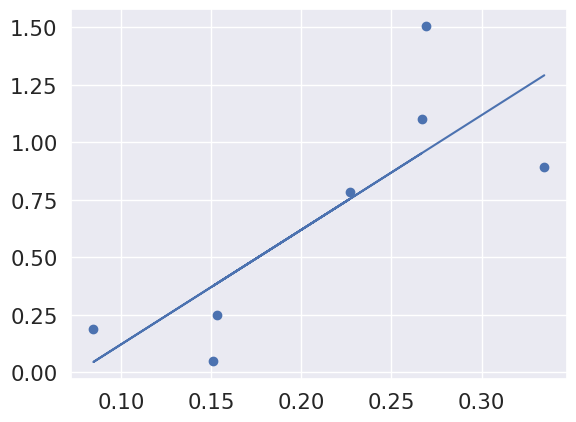}
    \caption{Scatter plots for presidential candidates, speeches (left) and speakers (right).}
    \label{fig:enter-label}
\end{figure}

\begin{figure}
    \includegraphics[width=0.5\linewidth]{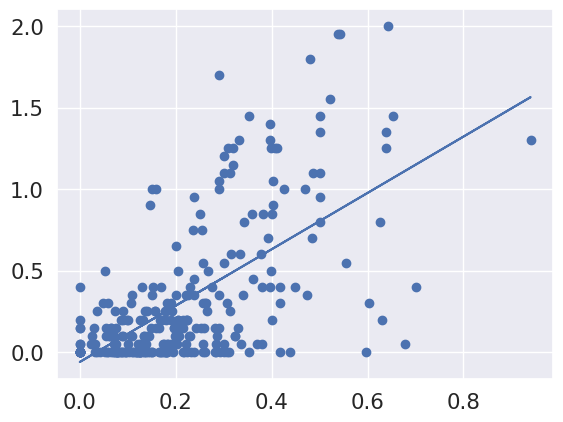}
    \includegraphics[width=0.5\linewidth]{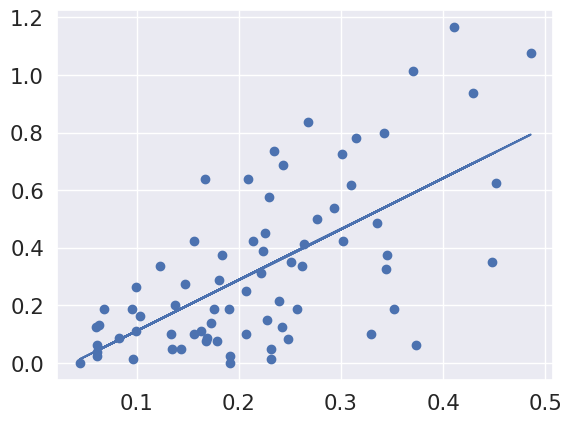}
    \caption{Scatter plots for governors with 2010-2018 terms, speeches (left) and speakers (right)}
    \label{fig:enter-label}
\end{figure}

\begin{figure}

    \includegraphics[width=0.5\linewidth]{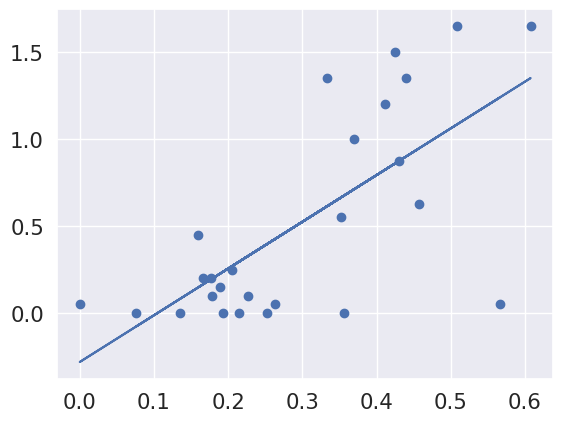}
    \includegraphics[width=0.5\linewidth]{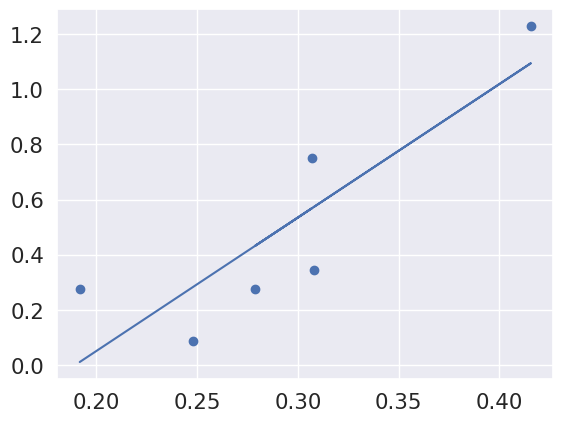}
    \caption{Scatter plots for governors with 2018-2022 terms, speeches (left) and speakers (right)}
    \label{fig:enter-label}
\end{figure}

We decided not to use these continuous variables because of the imbalance of the data. With many more non-populist than populist politicians, most data by far was in the 0-0.5 range, indicating very low populism scores. In addition, there were only few very high scores (of 1,5 or higher), with only 14 of the 358 speeches obtaining such a score. With so few high scores, we were concerned that the results would not be reliable at that range. Thus, we decided to group the data into a binary variable, with at 0,5 as the cutoff.

\begin{figure}
    \centering
    \includegraphics[width=1\linewidth]{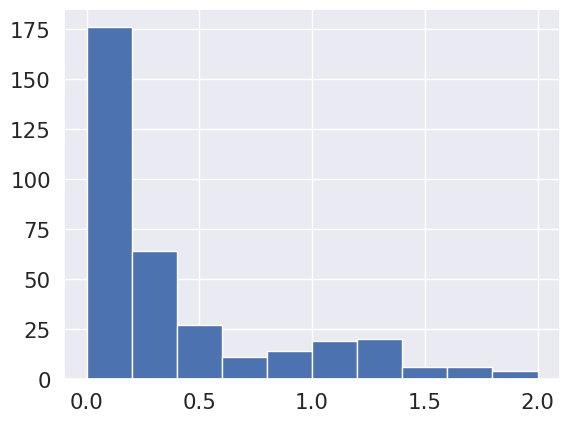}
    \caption{Histogram showing the distribution of scores in all data}
    \label{fig:enter-label}
\end{figure}

\subsection{Example sentences}

Below are 10 randomly selected (via the 'random sample' function in Python) sentences from each category and dataset. It is evident that most sentences are indicative of the category they are in, while some lose their significance when being split to the sentence level.

\textbf{Populist, presidential candidates}:

"We will all come together to say loudly, and clearly that the government of our great nation belongs to all of us, not just a few wealthy campaign contributors",

"And the only thing I say is hopefully, hopefully, our patriotic movement will overcome this terrible deception",

"The press can’t write the kind of things they write, which are lies, lies, lies",

"But there are still too many powerful interests fighting to protect their own profits and privileges at the expense of everyone else",

"They're going to lose their campaign funds from the Koch brothers and the fossil fuel (industry)",

"A Trump Administration will end the government corruption",

"A lot of those deals are made because the politicians aren’t so stupid",

"The Clintons ripped off the people of Haiti as they were suffering and dying after the earthquake",

"But the people are so incredible that I've been saying make America great again and it's going to be greater than ever before",

"Her donors will make sure of it"

\textbf{Populist, governors}:

"Too many people are falling behind financially, even as they work harder and harder",

"They have no political desire to reduce rates for Mainers",

"this election is about values and it's very clear that the Democratic Party has the values of the people",

"He might be the only person in Illinois who doesn't think corruption is a problem",

"I beg the people of Newark to repel such conversation, and to stand up for your right to choose",

"There’s now 400,000 dead",

"He said in Wyoming, where you find one blade of grass you leave two",

"But here’s the deal: the heroes at the Alamo, they are long gone",

"Turning up the heat",

"Bernard and his whole family, they’ve already voted".

\textbf{Pluralist, presidential candidates}:

"Everyone should be respected by the law, and everyone should respect the law",

"We stand up for religious and ethnic minorities, for women, for people with disabilities and we comport ourselves with honor",

"And if you agree – whether you’re a Democrat, Republican or independent – I hope you’ll join us",

"You know, we need to come together around all of the issues that are important to everybody and as I said in the beginning, whatever issue you care about, it's going to be on the ballot",

"I want to congratulate Senator Sanders on a hard fought race here",

"I do this for my family, of course, for my sweet family, for my neighbors, for my friends of many, many, many years, many of whom are working with me today—30, 40 years later",

"Jobs that provide dignity and a future",

"Whether you care about new, good jobs with a rising incomes or you care about better education or you care about what we can do to get the cost of prescription drugs down",

"We had lost 350,000 jobs",

"We have to rebuild our bridges, our airports, our hospitals in this country",

\textbf{Pluralist, governors}:

"Instead of simply demonizing the other political party, we bring Democrats, Republicans, independents together, instead of pitting business against labor or urban folks against rural folks or one side of the state against the other",

"Let's get it solved on a national level",

"That's what I would like to see",
 
"We can make it a bipartisan reality here in North Carolina",
 
"They got rings in the ears and nose and other places that I don't even know about",

"And shortly after that he came to Cheyenne and announced more  coal leases",
 
"And during the last seven years, California has reduced the unemployment rate from 12",
 
"There always be people within your own party, both of them, who will cheer you for being obstructionist and taking shots and playing the same old political games of the past",
 
"The governor refuses to accept our federal tax dollars back to take care of our working for our hospitals, while our dollars go to all the 30 states that didn't expand their programmes, taking care of their working for and reimbursing their hospitals, while ours are without and yes, 14 of those 30 states are led by Republican governors".

\textbf{Neutral, governors}:

"Even if they are not supporting me now, I support them",

"You are in front of 50, 60, 70, 80 million people",

"And let's finally deliver something long overdo, equal pay for women in this economy",
 
"We’re going to get them care and we’re going to pay for their – that care",
 
"So we are going to invest for our young people in education, in jobs, not jails or incarceration",
 
"Let us never forget, Democrats and progressives win when voter turnout is high",

"It's tough",
 
"We want to be leading the world in production, exports and manufacturing",

"My family were builders of a different kind",

"He told his men to stay back",
 
 "That still sounds good to me".

\textbf{Neutral, presidential candidates}:

"We shouldn’t walk away from high expectations, it is time to deliver on high expectations",

"I prosecuted Child Abuse and Child homicide cases",

"Are we safe",

"I mean about a third of any submarine assembled, as manufactured here in Connecticut, but they're not all assembled here",

"In rural Colorado, only 7 in 10 households have access",

"So, it's a technology challenge, it's a value challenge, it's a local challenge, and it's a global challenge",

"Governor Castro came to this country as a poor immigrant",

"Look at me now",

"And as was said earlier, I think by Santa Cruz, it's not about hunting, which it does protect, and I'm a hunter, that's a good thing",

"He promised eight years ago he'd cut people's taxes, but he didn't lower taxes for any".

\subsection{Speeches used for error analysis}

\subsubsection{Jindall}

Thank y'all. Thank y'all very much for that very generous reception is so great to be back here in Iowa with you. Now, even before I start, you know, it's great to be in a church. I know, coming to church makes some politicians a little nervous, but I start sweating. I don't come into church very often. I've been to nearly every church in Louisiana. I'm going to start off with just an old joke. I'm gonna warn you. It is an old joke. I'm going to tell it to you anyway.

I was in church in northwest Louisiana, and there was a pastor that was going on and on and on speaking longer than his congregation was listening.

Given how long you've been sitting, maybe you can sympathise a little just kind of gration. Anyway, at some point, the pasture is talking about a dear friend of his who had served in the Army who would who had just passed and he talked about how he had lost his friend in the service. All of a sudden, a woman woke up in the congregation and said, Pastor, what service was that the 930 or the 11 o'clock service?

Our guys an awesome God, amen.

I'm so excited to be here to talk about two of my favourite things, faith and freedom. Let's start with faith. You know, as an elected official, as a candidate, you get asked a lot of questions. today. I've talked to the media all day long. I will media national media, they'll ask you questions about schools, taxes, education, all kinds of things. I remember the easiest question I was ever asked. It was my very first time running for office. It was the last debate of that election. We were on live statewide TV. There were two candidates left. It was the last question. I had to answer the question first, they asked us this. They said what was the single most important moment in your life?

Now remind you this is live TV. You can't call a friend. There's no timeouts. You can't ask anybody for advice. You got about 30 seconds to answer. You know, I've been blessed in my life. married to a beautiful woman I first met in high school. At that time, I had one beautiful child, a little baby girl, we were expecting our second child. I'm going to talk about my wedding day, I could have talked about the first moments, I held my baby in my hands.

Instead, this is what I said.

I said the single most important moment in my life, is the moment that I found Jesus Christ.

   2:36  
I was being honest, I would say it was a moment he found me. It's not like he was the one that was lost. I'm the one that was lost.

   2:43  
I want to share with you just briefly how I came to my faith, I had an overnight epiphany that only took about seven years to happen.

   2:51  
Started with a best friend that came up to me on a school bus and said, Bobby, I'm gonna miss you when you're not in heaven.

   2:59  
Now, I don't encourage me to go and tell your friends stuff like when you say things like that people think you're crazy.

   3:05  
My best friend planted a seed of the gospel in my life. Several months later, he gave me a Christmas present I did not want. He gave me my very first copy of the Bible. But he was smart. He gave me a Bible with my name printed on gold letters on the front cover.

   3:25  
He knew I couldn't return it, I couldn't exchange it, I had to keep it.

   3:31  
God use that experience to plant another seed in my life. I didn't open that Bible till many years later when my grandfather suddenly died.

   3:41  
And I'd love to tell you, I open those pages. And the truth just jumped out at me. But it wasn't that easy. I was confused by what I read. But God use even that experience with Scripture to plant a seed. And there are many people that planted a seed in my life. It was finally years later, when my best friend that had given me this Bible invited me to come listen to him sing in his church, on the LSU campus.

   4:04  
And in the middle of that music, they showed a black and white film with an actor playing Jesus being crucified.

   4:10  
And I don't know why. But after all those years, God chose that moment to Hit me harder than I've ever been hit before.

   4:18  
Nothing special about that movie, nothing special about that actor, except that God chose that moment. When I saw the actor playing Jesus being crucified, it just hit me.

   4:27  
If that is really the Son of God,

   4:31  
and if he's really up there on that cross dying because of me,

   4:35  
dying because of my sins, auto mean dying for all of humanity. That's too easy. I mean, he is up there because of bobby jindal. And what I have done and what I have failed to do.

   4:47  
How arrogant for me to do anything but get them on knees and worship him.

   4:52  
I have made it so complicated. It was as simple as that. youth pastor handed out cards and said if you've not accepted Jesus Christ,

   5:00  
As your personal Lord and Saviour, we want to get to know you better. I filled out one of those cards, and we started meeting every month, and then every week and then every day. And we started reading scripture together, we started praying together. And as we read the New Testament, as we read the Gospels, it was like the Bible have been written just for me. The words were jumping off the page. It was like Jesus was talking me from 2000 years ago. And that's why it was so important in that moment. And it was so easy for me to say in that moment, the single most important moment in my life, that's easy. It's the moment that I found Jesus Christ is the moment he found me. Now, the one lesson I hope you take away from my testimony we all have our own testimonies, is the importance of planting seeds of the gospel, you may change some of these life for all of eternity, and not even realise it.

   5:55  
Now, I've talked a little bit about my faith, I want to talk to you now I want to shift gears a little bit and talk a little bit about freedom, and what's going on in our country today. You know, there's a lot that President Obama is doing that worries me about our country.

   6:08  
\$18 trillion of debt, an EPA that's trying to smother our economy, Obamacare, we have bureaucrats between doctors and their patients. You've now got the Federal Department of Education trying to force Common Core in our classrooms. You've got a foreign policy where we refused to stand with Israel, and we're about to allow Iran to become a nuclear power. You've got a president who won't even name the opponents we face in radical Islamic terrorists. But don't worry, he's keeping us safe from those mediaeval Christians that might come back.

   6:43  
So much of this can be undone. So much of this can be undone when we get a conservative in the White House. But the thing that worries me the most, is what the President is doing to redefine the idea of the American dream, because the changes he's making in our culture are going to be far more difficult to change back. When you listen to this president, he tries to divide us. He tries to divide us by race, by geography, by gender, by age by income. When he talks about the American dream, it's about government spending and borrowing. It's about government dependence.

   7:18  
That's not the government dream. That is the European nightmare. And the reason I feel so strongly about this is my parents, they have lived the American dream. My dad's one of nine. So the only one that got passed, the fifth grade in his family grew up in a house without electricity without running water. I know because we heard these stories every single day growing up.

   7:44  
Good luck trying to get an allowance from a father like that.

   7:49  
But here's the amazing thing about my parents, over 40 years ago, the very first time they'd ever gotten on a plane, they flew halfway across the world to come to Baton Rouge, Louisiana.

   8:03  
You know, I tried to tell my kids, they don't understand this. You couldn't Google America. Back then there was no online there were no computers at home. They had never even met somebody who had visited Louisiana and can come back and tell them what it was like.

   8:17  
They were coming to an idea as much as they were coming to a place. They were coming to the idea of freedom and opportunity. They were coming to the American dream. And you know what they caught the American dream. They landed in Baton Rouge my mom went to LSU. My dad got his first job by calling through the Yellow Pages, called company after company after company. Finally, a guy from a railroad company told him he could start on Monday morning.

   8:45  
I love what my dad tells us new boss has even met him yet. says that's great. So now Look, I don't have a car. I don't have a driver's licence. Does this new boss you're gonna have to pick me up on the way to work Monday morning.

   8:59  
The guy was so impressed by his enthusiasm, the work he did just that.

   9:03  
Six months later I was born. I was what you would politely call a pre existing condition.

   9:13  
That's what I've been called worse. It's okay.

   9:16  
All that meant was my parent's insurance didn't cover me and I love what my dad did next. I was born in a woman's Hospital in Baton Rouge where two of our three children were born. My dad went to the doctor after I was born.

   9:28  
said look, insurance doesn't cover our child's birth. So I'm going to send you a check every month until I pay this bill in full.

   9:44  
No paperwork, no contracts, no government programme. Just two guys shaking hands in the hospital.

   9:52  
You know it was a simpler time back then. That's what people used to do.

   9:56  
I asked my dad How did that work today.

   10:00  
Do you pay for a baby on layaway?

   10:03  
I mean, if you skip a payment, can they take the baby back? What do they do?

   10:08  
My father's made it very clear. He said, Son, you were such a bad baby, we would have sent you back. That was an option.

   10:15  
He said, Don't worry, you're paid for Don't worry, nobody's coming to take you. The reason I tell you this, Mark Twain said, the older we get the smarter our parents become.

   10:25  
I don't know about you, I find myself turning into my parents more and more every day. I say things I swore I would never say one of the things my dad used to tell my brother and me is that I'm not leaving a famous last name or an inheritance. But I'll make sure you get a great education. Because in this country, there's no limit on what you can accomplish. But it also tells us he said every day You should thank God, that you are blessed to be born in the greatest country in the history of the world, the United States of America.

   11:02  
I'm about you. I want my children and one day my grandchildren be able to say that same prayer. And I worry about the assault on freedom, the redefinition the American dream, and we're seeing it right now. We're seeing an unprecedented assault on our religious liberty rights in the United States of America.

   11:25  
A little over a year ago in February at the reagan library, I gave a speech outlining the upcoming assault on religious liberty. It's no longer upcoming folks, it is here. And it didn't start in Indiana and Arkansas. It didn't start in the Hobby Lobby case where the Obama administration wanted to force the green family to use their own money to pay for abortifacients to pay for abortions. The violated their sincerely held religious beliefs. And it certainly it didn't start. It didn't start when the left got so mad at Phil Robertson for saying things that they disagree with. They tried to get Duck Dynasty cancelled on a\&e. And it certainly didn't start this week, when Hillary Clinton stood up in New York and said, those of us that are pro life,

   12:10  
those of us that are pro life need to have our religious beliefs changed.

   12:16  
Now listen to what she said we need to have our religious beliefs changed.

   12:21  
I don't know how she proposes to do that. She didn't suggest whether we should go to reeducation camps or how she intends to do that. But I've got news for her. My religious beliefs are not between me and Hillary Clinton. My religious beliefs are between me and God. And we're not.

   12:46  
We're not changing our religious beliefs simply because they upset Hillary Clinton.

   12:51  
Now we saw something very unusual in Indiana, we saw corporate america team up with the radical left to come after our religious liberty rights. Corporate America needs to be careful. The same radical left that doesn't want us to have religious liberty rights does not want us to have economic liberty rights. The reality is the same radical left that doesn't want us to have religious liberty rights, wants to tax and regulate these companies out of existence. I think profit is a dirty word. Corporate America needs to be careful. I know that they think they succeeded in bullying those leaders in Indiana, but I've gotten used for them. We've got legislation in Louisiana, we've already got a religious freedom act, we've got legislation this session, to protect people of faith and of conscience who hold a traditional view of marriage. And they might as well save their breath because corporate America is not gonna bully the governor of Louisiana when it comes to religious liberty.

   13:53  
And they need to understand there is no freedom of speech or freedom of association without religious liberty. I reject this notion that in America, it is impossible have religious liberty and also to get rid of discrimination. The reality is we can do both. And the reality is this. The real discrimination that is being faced today are Christians, individuals, families and business owners that shouldn't have to choose between operating their businesses, and following their conscience their traditional views their religious beliefs. When Hillary Clinton says we have the freedom of religious expression, that's not religious liberty. All she means is for a couple of hours a week we can say what we want in church. Religious Liberty means being able to live our lives 24 hours a day, seven days a week according to our faith according to our conscience, according to our beliefs. Now, this fight is bigger than marriage. I believe in traditional marriage between a man and a woman and unlike President Obama and Secretary Clinton, the governor of Louisiana, his views my views. They're not evolving with the times. They're not based on phone numbers.

   15:00  
This fight is bigger than even marriage though. This fight is about the definition of liberty and freedom in the United States. The left thinks we're not smart enough to have Second Amendment rights, we're not smart enough to have school choice. We're not smart enough to drink a big gulp in New York, we're not smart enough to we're not smart enough to buy our own health insurance. Now, they don't think we're smart enough to live our own Lodge. Well, I've got a message to you at one time, the left the media leaked the Hollywood elite, they used to believe in tolerance. And they still do they tolerate everybody except those who happen to disagree with them.

   15:34  
It's no coincidence, the one group they do want to discriminate against or evangelical Christians with traditional views. So the only group, I think it's okay to discriminate against, I've got a message for the left. And I'll say it simply and slowly, so they'll understand it. It's not real complicated.

   15:49  
You don't have to go to Harvard Law to understand this could have saved the president three years a lot of tuition money.

   15:54  
He's asked for his tuition money back, I'm not sure what he learned there anyway.

   15:59  
They certainly didn't teach the constitution when he was there.

   16:09  
Here's my message to Hollywood to the media leave. The United States of America did not create religious liberty, religious liberty created the United States of America. And is the reason why

   16:44  
I want to close with this following observation. I told you about the American dream. And I told you, all of my children and your children and grandchildren to be able to pray the same prayer that we are blessed to be born and raised in the greatest country in the history of the world. But I will say this, we've had a president for over six years now is trying to divide us. I do believe our best days are ahead of us. But we've got to win this election. We've got to beat Hillary and I'm one of those that says, We don't just need to incremental change in DC. We don't just need to like any republican we need big change in Washington, DC to get our country back on the right track.

   17:25  
But my final thought is this my parents, they loved India.

   17:30  
But when they left India, they were coming to America to be Americans. If they wanted to be Indians, they would have stayed in India. If they wanted to raise their children as Indian Americans, or as in New Jersey, anything else they wouldn't have come here. My parents came here to raise Americans. I don't know about you.

   17:56  
I'm tired of the hyphenated Americans. We're not African Americans. We're not Indian Americans are not Asian Americans. We need to stop dividing ourselves. We are all Americans, Americans.

   18:18  
It used to be common sense to call America the great melting pot. It used to be common sense to say that those that want to come here should want to be Americans. I'm tired of a commander in chief. I'm tired of a president that seeks to delight in criticising America apologising for America. Our founding fathers got it right.

   18:35  
Our founding fathers got it right. No external enemy can beat us but we can be beaten if we were divided within the purpose of our governance, not to create rights is to secure our God given rights. This is our time. This is our time, our four youth presidents that every generation has to renew for itself, our principles of freedom. Previous generations have spilt blood and treasure to give us this inheritance. This is our time. My parents knew what they were doing when they came here over 40 years ago in search of the American dream. Let's make sure the next generation gets a chance to live that American dream as well. God bless y'all. Thank y'all very much. Welcome, si like so great to be with y'all.

\subsubsection{Clinton}

Hello, Dade City!

I am so excited about being here. Thank you, all for this really warm, wonderful welcome. And I thank, on behalf of all of us I want to thank Alicia Machado for that introduction and for sharing her story with us. Alicia will be voting for the very first time in this election, and I am very grateful for her support. I'm also delighted that we're joined today by your great senator Bill Nelson and his wife Grace.

I also want to thank Representative Amanda Hickman from the Florida State House, Michael Cox, chairman of the Greater Pasco Chamber of Commerce, Michael Ledbetter, who is the Pasco County Democratic executive committee chair and his wife Beverly. And thanks to retired Colonel Wilson Elton [inaudible] for his pledge of allegiance. And thanks to all the elected officials here. And I especially want to thank all of you for not just joining us but helping us to get out the vote in this last week!

It's almost hard to believe, isn't it? There are only seven days left in this election.

So are you ready to vote?

Are you ready to volunteer these last seven days?

I hope too that you ready to elect Patrick Murphy to the United States Senate! Patrick, who's been in Congress, will be an independent voice for Florida families. And I am so excited to think about what he could do, because unlike his opponent, he's never been afraid to stand up to Donald Trump.

Now somebody asked me the other day, 'Why do you keep coming back to Florida?' Just look around folks. I mean it is a beautiful, and I have lots of friends, but it's also really important in this election. Florida can decide who our next president is, which will affect the nation and the world. And I want to make sure that every voter in Florida spends these next seven days thinking about what's at stake in this election. Because honestly I believe this may be the most important election of our lifetimes.

One week from today, we will be choosing our next president and commander-in-chief of the United States. I don't think the choice could be any clearer. I have spent my career fighting for children and families. I have served in the United States Senate, served on the Senate Armed Services Committee. I was in the Situation Room when we brought Osama Bin Laden to justice. I represented you as your secretary of state, going to 112 countries, negotiating with friend and foe alike. I am ready to serve if you give me the great honor of being your president.

Now, I do have to say that I think that stands in contrast to my opponent. And maybe for you if you think about all of the issues that separate Donald Trump from me, it could be his dangerous statements about nuclear weapons.

When a journalist told Donald Trump that people were worried about how casually he talks about using nuclear weapons, he said, 'Well then why are we making them?'

And I have to tell you yesterday I was in Ohio, and I was introduced by a gentleman who was one of our officers in charge of launching nuclear weapons if the order ever came from our president. And what he told the crowd — it was a big crowd at Kent State University — I think every voter should hear. Basically — his name was Bruce Blair — he said that, having had the responsibility of sitting in a bunker, being responsible for the codes and the keys to our nuclear weapons, he knows that when a president makes the order to launch a nuclear attack, there is no appeal. There is no veto. And that is why he has joined with dozens of former Air Force officers to send a letter to say, 'We need a president with the temperament, the steadiness, the calmness to be in charge of nuclear weapons.' And therefore they cannot support Donald Trump, because he does not have the temperament to be our commander-in-chief and handle those responsibilities.

So when you think about voting, early this week, voting next Tuesday, responsibility for our nuclear weapons is on the ballot.

So is immigration. Do we want to round up millions of people who are here working, raising their families, as he has suggested he will do? I don't think so. I think what we want is to bring them out of the shadows so that they can't be exploited by employers like Donald Trump, who refused to hire Americans and hired undocumented workers so he could pay them less. I don't think that's right.

Now maybe for some of you, what Donald Trump said about prisoners of war will be enough reason to vote against him, somebody who questioned the patriotism and the service of John McCain because he was a prison of war. We need a Commander in Chief who respects the service and sacrifice of the men and women who wear our uniform.

Or maybe for some of you, it's what he said about a judge born in Indiana, who just happened to be assigned the case brought by people who were defrauded by the phony Trump University. And so Trump said, 'Well, we can't trust him because his parents were born in Mexico.' Now Paul Ryan, the Republican Speaker of the House, called — well, but wait a minute, — he called what Trump said about that distinguished federal judge, 'The definition of a racist comment.'

And then Trump went on to attack a Gold Star family whose son, Captain Khan, died defending our country, simply because that family was Muslim.

And then let's not forget, Trump spent years, years insisting that President Obama was not born in the United States, even after the birth certificate was produced. Honestly if this were something new, I think we'd all be asking ourselves, 'Well, what does he have against President Obama?' Or what does he have against me? But this is not new. I know I'm reaching out to Republicans and Independents as well as Democrats because I want to be the president for all Americans.

And here's what I want you to tell. I want you to tell your Republican friends, in 1987 Donald Trump took out a Donald Trump took out a \$100,000 ad in the New York Times to criticize President Reagan. He said, 'Our leaders are the laughing stock of the world.' So this is a man who thinks that he is better than President Reagan, better than President Obama – literally better than anybody, I guess. And, when you think about it, what he said at the convention, 'I alone can fix it,' runs counter to who we are as Americans. We work together.

So there are many reasons why I think it is fair to conclude that Donald Trump is unqualified and unfit to be president. But today, I want to just spend a few minutes focusing in particular about what he has said and what he has done to woman and girls because – Any of you see the debates? I stood next to Donald for four and a half hours during those three debates, proving conclusively I have the stamina to be President of the United States. And during those debates, Donald always used to say, 'What have you been doing for 30 years?' And I always found that kind of odd, because he could Google it and find out. And so I've been a lawyer, and I've been a first lady, and I've been a senator, and I've been Secretary of State, and I've been a wife and a mother and a grandmother and a friend and a churchgoer, and for my entire life I've been a woman.

And when I think about what we now know about Donald Trump and what he's been doing for 30 years, he sure has spent a lot of time demeaning, degrading, insulting, and assaulting women. And, I've got to tell you, some of what we've learned – some of this stuff is very upsetting. I would, frankly, rather be here talking about nearly anything else, like how we're going to create good jobs and get the economy working for everybody, not just those at the top. How we're going to make college affordable for every single family, because I have a plan that if your family makes less than \$125,000, you will not pay tuition to go to a state college or university. And if you're above that, it will be debt free, and we will help you pay back the debt you already have so you can get out from under it.

But I can't just talk about all of the good things we want to do, because people are making up their minds. This is a consequential choice, so we've got to talk about something that, frankly, is painful, because it matters. We can't just wish it away. And a lot of his supporters don't like to hear this. I don't blame them. If I were supporting him, I wouldn't want to hear it either, to be honest. But I've got to tell you, I learned way back in elementary school and I learned it in Sunday school: it's not okay to insult people. It's not okay.

And look at what he does. He calls women 'ugly,' 'disgusting,' 'nasty' all the time. He calls women 'pigs,' rates bodies on a scale from 1 to 10. We just heard from Alicia. She was Miss Universe when Donald Trump owned the pageant. Well, he said she put on some weight and it made him angry, so he called her 'Miss Piggy.' He called her 'Miss Housekeeping' because she's a beautiful Latina. He brought a bunch of reporters to a gym to watch him order her around to exercise. Now, he also said, 'This is somebody who likes to eat.' Well, I have to say: who doesn't like to eat? I mean, really, can we just stop for a minute and reflect on the absurdity of Donald Trump finding fault with Miss Universe?

But you've got to ask: why does he do these things? Who acts like this? And, I'll tell you who: a bully, that's who. And thankfully, Alicia refuses to let such a small person get her down. She knows that Donald Trump doesn't get to decide her value in her eyes and in the eyes of her family and her friends.

But what about our girls? What about girls watching all this? What happens to their confidence, their sense of self-worth? If you've got a daughter, a granddaughter, a sister, a mother, a wife, a good friend, someone like this becoming president who insults more than half the population of the United States of America? And what about our boys? This is not someone we want them looking up to.

Not so long ago, I was the mother of a teenage girl. And every day I tried to make sure that she knew she was smart and she was capable, and I'm doing the same thing with my granddaughter and my grandson. Because, let's be honest here, the world has a way of telling our girls exactly the opposite – they don't look right, they're not thin enough, they don't act right, no one will like them unless they change their clothes or straighten their hair or stop being bossy, or whatever the criticism might be. Right?

And I remember, when Chelsea was a teenager, I would wait on the second floor of the White House in the long hall that's there for her to come home from school. And sometimes, we only had a few minutes together. But before she'd run off to talk to her friends or do whatever she was planning to, I would figure out how was the day. We'd talk about what was on her mind. And that time together was really valuable to me as a parent, because we parents, and I know there are a lot of parents here today, we work hard to give our kids a sense of confidence to send them out into the world believing in their own value, and it really is important that we don't let anybody take that away from them. And we want them to be strong, and we don't want them to feel bad about themselves. And we've got to work hard to make sure our boys, just like our girls, have that same sense of positive energy: not negative, not tearing people down – lifting people up and respecting women.

In fact, all of us should respect each other in our country. So, when I look at my granddaughter and my grandson, I am on the same mission. I want them to know they're love, they're cared for, they're respected. I want them to develop a good work ethic. I believe in hard work, and I want them to go out and prove themselves in the world. But that's what I want for every child. I have spent my life doing everything I could to help kids and families. It will be the mission of my presidency. I will get up every day in the White House trying to figure out: how can we knock down the barriers, overcome the challenges so that people living here in Pasco County can get ahead and stay ahead?

So I know there's work to be done. We can't do it with just words alone. We've got to do more to stop treatment about women being somehow objectified. And, oh my gosh, when we heard that tape and we heard what he does to women. I'm not going to repeat it. But you know what Donald Trump was bragging about: grabbing women, mistreating women. And I have to tell you, since that tape came out, twelve women have come to say that, 'What he said on that tape is what he did to me.'

And then we heard his response. What he does at his rallies is to go after those women all over again, insulting them. He said he couldn't possibly have said those things because the women weren't attractive enough to assault. 'Look at her,' he said about one woman. 'I don't think so.' About another, he said, 'She wouldn't be my first choice.'

He's also on tape bragging with the radio personality Howard Stern about how he used to go backstage at beauty pageants to barge in on the women while they were getting dressed. He said he did that — he said he did that to 'inspect' them. That was his word – and he said, 'I sort of get away with things like that.' And sure enough, contestants have come forward to say, 'Yes, that's exactly what he did to us.' Now, as bad as that is, he didn't just do it at the Miss USA pageant or the Miss Universe pageant. He's also been accused of doing it at the Miss Teen USA pageant. Contestants say that Donald Trump came in to look at them when they were changing. Some of them were just 15 years old.

We cannot hide from this. We've got to be willing to face it. This man wants to be president of the United States of America and our First Lady, Michelle Obama, spoke for many of us when she said Donald Trump's words have shaken her to her core.

For a lot of women who have gone through something like this in your lives, that's brought back painful memories. And for men who would never, ever talk or act like Donald, it's been shocking to see this. He tried to explain it away as 'locker room talk,' well, I'll tell you what, a lot of professional athletes stood up and said, 'Not in our locker rooms. That does not happen.'

Well, I guess the bottom line is he thinks belittling women makes him a bigger man.

And I don't think there's a woman anywhere who doesn't know what that feels like.

He doesn't look at us and see full human beings, with our dreams and purposes, our own capabilities. And he has shown us that clearly throughout this campaign.

Well, he's very wrong. He is wrong about both the women and the men of this country. He has shown us who he is. Let us on Tuesday show him who we are. We can stand up for what we believe, what we want for our children and grandchildren, what we know is right. And you can go down the list of everything he has said. He doesn't believe in equal pay, he thinks pregnancy is an inconvenience, he won't raise the minimum wage, and he said if he comes home and dinner is not on the table, he gets angry. Instead of supporting women who are out there supporting their families, he wants to make it even harder.

Well, I have a message for him. We're going to fight for affordable childcare. We're going to fight for equal pay for all people. We're going to fight for paid family leave, we're going to fight to make sure everybody gets raising wages in America and that's important because so many people still struggling, still working hard, raising a family, having a hard time getting all of their bills paid. And that's especially true for minimum wage workers — who — two-thirds of whom are women, so we've got work to do my friends. I'm excited because I know we can do this and I will stand up and I will fight for you, I will work for you, I will give my heart to this mission of making our country all it should be. Because instead of Donald Trump's dangerous and divisive vision, mine is positive, optimistic, hopeful and unifying.

But I can't do any of this without your help. Early voting has already begun. Almost 26 millions Americans have already voted and that includes four million right here in Florida. Americans are voting for the kind of better future we can make together. And voting on all the issues that they care about. It may be my name and Donald's name on the ballot but everything you care about, our security, our economy, bringing our country together, the environment, clean water and clean air!

Here in Dade City, you can vote every day from now until November 5th, from 7am to 7pm.

In fact, right now, the East Pasco Government Center is just a few minutes' drive away. You can go there right after this rally and vote. Because, we need everybody to stand up in this election.

And if you have a mail-in ballot at home, send it in when you get home today.

Don't wait to send it back! Talk friends and family and neighbors and co-workers.

Donald Trump's strategy is pretty simple– he said he wants to suppress young people from voting women from voting, people of color from voting, that is a lot of people. By showing up with the biggest turnout ever we will show him and everybody a message that that is not who we are.

Now I know here in this county and this larger region, you probably know some people who are going to vote for Trump. And here is what I want to ask you, I want you to talk with them. Ask them what they care about. Ask them what kind of future they want for their economy. Because Donald Trump's economic plan is slashing taxes on the wealthy and big corporations. I have said that I want the wealthy to pay their fair share and I will not raise taxes on anybody making less than \$250,000 a year.

So there could not be, no matter what you care about, a bigger difference between me and Donald Trump, and I hope that you will come out and volunteer for these last seven days.

We are signing up for volunteer shifts making calls, knocking on doors, it really matters, turn out matters.

Please go to hillaryclinton.com to see how you can get involved. Here is what I want to leave you with. You know, I feel like so many of you, our country is already great but we can make is greater. The main reason America is great is because America is good. We are a big hearted, generous people. Not a small minded people. We know that if everybody works together we will get farther together than if we separate people, we push them down, we engage in all the negativity that we have heard too much in this campaign. But sometimes, the fate of even the greatest nations lie in the balance. For America, this is one of those make or break elections. It really is in your hands.

I hope you will think about how you will feel the day after the election on November 9th. are we going to more forward together or are we going to move backwards by someone who wants to bully us? And I hope, I hope you will think about how together we can make a difference, I want to be your partner as well as your president, I want us to create the best possible future for our children and our grandchildren. That is what I will work my heart out to do and I hope you will help me to make a better, fairer, stronger America. Where we prove once and for all that love trumps hate!

Thank you all."

\immediate\write18{texcount -sub=section \jobname.tex -out=\jobname.wcdetail}
\end{document}